\documentclass[11pt,a4paper]{article}
\usepackage{caption}
\usepackage{tabularx}
\usepackage{booktabs}
\usepackage{graphicx}
\usepackage{amsmath}
\usepackage{amssymb}
\usepackage{subfigure}
\usepackage{algorithm,algorithmic}
\usepackage{multirow}
\usepackage[hyperref]{emnlp2020}
\usepackage{times}
\usepackage{latexsym}

\usepackage{microtype}

\aclfinalcopy 

\title{Computer Assisted Translation with Neural Quality Estimation and Automatic Post-Editing}

\author{Jiayi Wang\thanks{~~indicates equal contribution.}, Ke Wang$^*$, Niyu Ge, Yangbing Shi, Yu Zhao, Kai Fan\thanks{~~indicates corresponding author.} \\
Alibaba Group Inc.\\
\texttt{\{joanne.wjy,moyu.wk,niyu.ge,taiwu.syb\}@alibaba-inc.com},\\ 
\texttt{kongyu@taobao.com}, \texttt{k.fan@alibaba-inc.com}
}

\date{}

\begin{document}
\maketitle
\begin{abstract}
With the advent of neural machine translation, there has been a marked shift towards leveraging and consuming the machine translation results. 
However, the gap between machine translation systems and human translators needs to be manually closed by post-editing. 
In this paper, we propose an end-to-end deep learning framework of 
the quality estimation and automatic post-editing of the machine translation output. 
Our goal is to provide error correction suggestions and to further relieve the burden of human translators through an interpretable model. 
To imitate the behavior of human translators, we design three efficient delegation modules -- quality estimation, generative post-editing, and atomic operation post-editing and construct a hierarchical model based on them. 
We examine this approach with the English--German dataset from WMT 2017 APE shared task and our experimental results can achieve the state-of-the-art performance. 
We also verify that the certified translators can significantly expedite their post-editing processing with our model in human evaluation.
\end{abstract}


\section{Introduction}


The explosive advances in the sequence to sequence model \cite{sutskever2014sequence,bahdanau2014neural,vaswani2017attention} enable the deep learning based neural machine translation (NMT) to approximate and even achieve the human parity in some specific language pairs and scenarios. 
Instead of translating from scratch by human translators, a new translation paradigm has emerged: computer assisted translation (CAT) system, which includes the machine translation and human post-editing. 
The post-editing is the process whereby humans amend machine-generated translations to achieve an acceptable final product. 
Practically, the estimated average translation time can be reduced by 17.4\% (from 1957.4 to 1617.7 seconds per text) \cite{laubli2013assessing}.

However, utilizing NMT poses two key challenges. 
First, the neural machine translation quality still continues to vary a great deal across different domains or genres, more or less in proportion to the availability of paralleled training corpora. 
Second, the zero tolerance policy is a common choice in the vast majority of important applications. 
For example, when business legal documents are translated, even a single incorrect word could bring serious financial or property losses. 
Therefore, the subsequent human post-editing is indispensable in situations like this. 
Unfortunately, while NMT systems saves time by providing the preliminary translations, the time spent on error corrections by humans \cite{laubli2013assessing} remains substantial to the extent that it offsets the efficiency gained by the NMT systems. 
In this paper, we explore automatic post-editing (APE) in the deep learning framework. 
Specifically, we adopt an imitation learning approach, where our model first screens the  translation candidates by quality prediction and then decides whether to post edit with the generation or the atomic operation method.

Starting with a wide range of features used in the CAT system, we carefully analyze the human post-editing results to narrow down our framework design into three key modules: quality estimation (QE), generative post-editing and atomic operation post-editing. 
These modules are tightly integrated into the transformer neural networks \cite{vaswani2017attention}.  
Our main innovation is a 
hierarchical model with two modular post-editing algorithms which are conditionally used based on a novel fine-grained quality estimation model.
For each machine translation, our 
model i) runs the QE model to predict the detailed token level errors, which will be further summarized as an overall quality score to decide whether the machine translation quality is high or not, and ii) conditional on the previous decision, employs the atomic operation post-editing algorithm on the high quality sentence or the generative model to rephrase the translation for the low one. 

We examine our approach on the public English--German dataset from WMT\footnote{\url{http://www.statmt.org/}} 2017 APE shared task. 
Our system outperforms the top ranked methods in both BLEU and TER metrics. 
In addition, following a standard human evaluation process aimed at achieving impartiality with respect to the efficiency of CAT system, we ask several certified translators to edit the machine translation outputs with or without our APE assistance. 
Evaluation results show that our system significantly improves translators' efficiency. 

\section{Related Work}
Our work relates to and builds on several intertwined threads of research in machine translation, including QE and APE. 
We briefly survey the traditional methods and differentiate our approach.

\subsection{Quality Estimation} 

Quality estimation is often a desired component for developing and deploying automatic language technologies, and has been extensively researched in machine translation \cite{barrault-etal-2019-findings}. 
Its purpose is to provide some metrics measuring the overall quality. 
The current state-of-the-art models mostly originated from the predictor-estimator framework \cite{kim2017predictor}, where a sequence-to-sequence model is pre-trained to extract sophisticated sequence features to be fed into a sequence level regression or classification network. 

\citet{tan2017neural} proposed the neural post-editing based quality estimation by streamlining together the traditional QE and APE models. 
Since our proposed QE module will eventually serve the APE module as well, we consider two modifications accordingly. 
First, we re-define the QE as a fine-grained multi-class problem, whose output indicates the number of tokens in four categories, missing / redundant / erroneous or kept tokens. 
A similar idea was initially proposed in \cite{gu2017non} to predict the number of copy occurrences in non-autoregressive neural machine translation. 
In this paper, we make significant extensions to include more categories. 
Secondly, we maximize our QE model performance with a novel conditional BERT architecture. 
Inspired by the masked language model objective in the encoder BERT \cite{devlin2019bert}, we introduce the training objective to the encoder-decoder framework by adapting the decoder to become a memory encoder, allowing us to pre-train the target language model similar to BERT but conditioned on the source language text. 

\begin{table}[t]
  \caption{Notation used in the model}
  \label{tab:notation}
  \begin{small}
  \begin{tabularx}{1.0\linewidth}{c|X}
    \hline
    Symbol & Definition\\
    \hline
    $\mathbf{s}$ & sentence in source language \\
    $\mathbf{m}$ & machine translated sentence in target language \\
    $\mathbf{t}$ & golden (reference) sentence in target language \\
    $\mathbf{e}$ & post-editing sentence in target language \\
    $s_i$ & the $i$-th token of $\mathbf{s}$, similar for $m_i,t_i,e_i$ \\
    $P_{\text{MT}}$ & the probabilistic model of machine translation \\
    $P_{\text{PE}}$ & the probabilistic model of post-editing \\
    $P_{\text{QE}}$ & the probabilistic model of quality estimation \\
    $\mathbb{I}_A$ & indicator function, $=1$ if $A$ is true, o.w. 0 \\
    $\tau$ & threshold to distinguish high/low quality translation \\
  \hline
\end{tabularx}
\end{small}
\end{table}

\subsection{Automatic Post-Editing}

Automatic Post Editing aims to improve the quality of an existing MT system by learning from human edited samples, converting ``translationese" output into natural text. 
The traditional APE is based on a round-trip translation loop to mimic errors similar to the ones produced by NMT and can achieve acceptable performance with large scale monolingual data only \cite{freitag2019ape}. 
However, the prevalent trend in this area prefers the dual-source encoder-decoder architecture with parallel data \cite{chatterjee2017multi,MSUEdin18APE,pal2018transformer,lopes-etal-2019-unbabels}, which obtained the best results in WMT competitions \cite{chatterjee2019findings}. 
The dual-source encoder encodes the source text and the machine translation output separately, and the decoder decodes the post-edited results. 
All these approaches encode each source independently and apply an auto-regressive decoder. 
They differ in their parameter sharing mechanisms. 

While our approach still employs the multi-source APE framework, but there are two fundamental differences. 
First, our APE module, as aforementioned above, is built on our re-designed QE model, with which the source and the machine translation are entangled by the encoder and memory-encoder QE module. 
Second, our decoder consists in a versatile architecture that can choose between the left to right auto-regressive generative model and the atomic-operation based paralleled model. 
It dynamically determines which model to engage at runtime. 
The parallelizable model was broadly explored in insertion- or deletion- based transformer \cite{chan2019kermit,stern2019insertion,gu2019levenshtein}, while our decoder supports more functional operations.

\section{Model and Objective}

\begin{figure}[!bhp]
  \centering
  	\includegraphics[width=\linewidth]{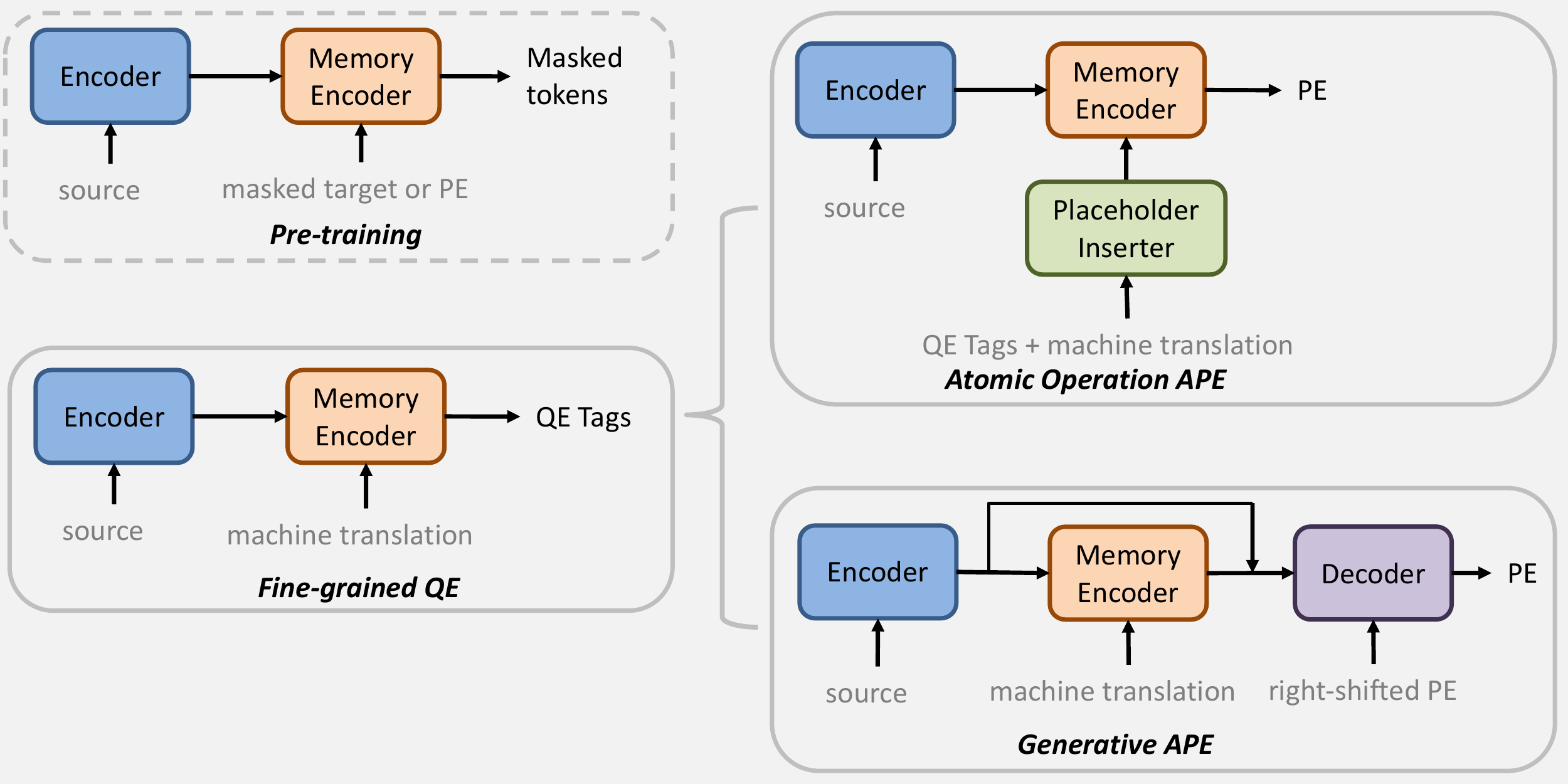}
  \caption{The overall pipeline. The QE model will output fine-grained metrics to the translation quality. Then, high quality machine translation will proceed with atomic APE model for minor fix, while the low quality machine translation will go through a generative APE model for completely rephrasing. Note that the model parameters are shared for three steps w.r.t. encoder and memory encoder. Detailed computational graph can refer to Figure \ref{fig:detail_model}.}
  \label{fig:model}
\end{figure}

In order to achieve the automatic post-editing goal, it is essential for the model to find the exact errors appearing in the machine translation and learn how to fix them. 
Breaking the problem into several subtasks, our proposed pipeline includes three major models as Figure~\ref{fig:model}. 
By skipping the pre-training temporarily, the first step is to investigate the fine-grained quality estimation model with respect to the source text and machine translated text. 
Its output will provide a fine-grained quality estimation of the machine translation. 
Based on the corresponding quality, an atomic APE or a generative APE model will be called for further processing. 

\subsection{Fine-Grained Quality Estimation}
\label{sec:qe}

\begin{table}[h]
\setlength{\tabcolsep}{2pt}
  \caption{Definition of QE Tags}
  \label{tab:qe_tag}
  \begin{small}
  \begin{tabularx}{1.0\linewidth}{c|c|c|c|c}
    \hline
    Label & $k > 1$ & $k = 1$ & $k = 0$ & $k = -1$ \\
    \hline 
    Definition & insert $k-1$ tokens & keep & delete & replace \\
  \hline
\end{tabularx}
\end{small}
\end{table}

As described in the related work, compared to traditional translation QE task in WMT\footnote{\url{http://www.statmt.org/wmt19/qe-task.html}}, our QE module is more fine-grained and is recast as a multi-class $\{-1,0,1,...,K\}$ sequence labeling problem. 
The definition of the integer labels is shown in Table~\ref{tab:qe_tag}. 
If $k<=1$, the label denotes one single token operation; otherwise, it means to insert $k-1$ extra tokens after the current one. 
The QE tag $\mathbf{q}$ for training pair $(\mathbf{m}, \mathbf{e})$ can be deterministically calculated by dynamic programming Algorithm~\ref{alg:qe_tag} in Appendix, which is basically a string matching algorithm. 
We define a conditionally independent sequence tagging model for the error prediction. 
\begin{equation}\label{eq:qe_tag}
P_{\text{QE}}(\mathbf{q}|\mathbf{s},\mathbf{m}) = \prod_i P_{\text{QE}}(q_i|\mathbf{s},\mathbf{m})
\end{equation}
A transformer based neural network is employed. 
We present a novel encoder-memory encoder framework with memory attention as shown in the decomposition of the following equation.
\begin{equation}\label{eq:qe_tag2}
\begin{split}
&P_{\text{QE}}(\mathbf{q}|\mathbf{s},\mathbf{m}) \\ \triangleq  ~&\text{Softmax}_{\text{QE}}(\text{Enc}^M(\mathbf{m}, \text{Enc}(\mathbf{s})))
\end{split}
\end{equation} 
where $\text{Enc}(\cdot)$ is the standard transformer encoder \cite{vaswani2017attention}, and $\text{Enc}^M(\cdot)$ is the memory encoder adapted from standard transformer decoder. 
It removed the future masking in the transformer decoder and use the last state as the output which contains contexts from both SRC and MT.

During inference, neither the ground truth of post-editing nor the golden translation reference is available. 
The fine-grained QE model can predict the human translation edit rate (HTER) $h$ through the inferred QE tags $\hat{\mathbf{q}}$. 
\begin{equation}\label{eq:hter}
\begin{split}
h &= \frac{\# \text{predicted edits}}{\text{predicted PE length}} \\
&= \frac{\sum_i \{\mathbb{I}_{\hat{q}_i < 1} + (\hat{q}_i - 1)\mathbb{I}_{\hat{q}_i >= 1} \} }{\sum_i |\hat{q}_i|}
\end{split}
\end{equation}
On the one hand, the overall metric $h$ can quantitate the quality of machine translation and determine which APE algorithm will be used. 
On the other hand, the detailed QE tags can theoretically guide the APE which atomic operation should be applied. 
Thus, the QE tagging and the atomic operation APE are simultaneously and iteratively trained, which will be elaborated in \ref{sec:aoape} and \ref{sec:alg}.

\begin{figure*}[t]
  \centering
  \includegraphics[width=0.85\textwidth]{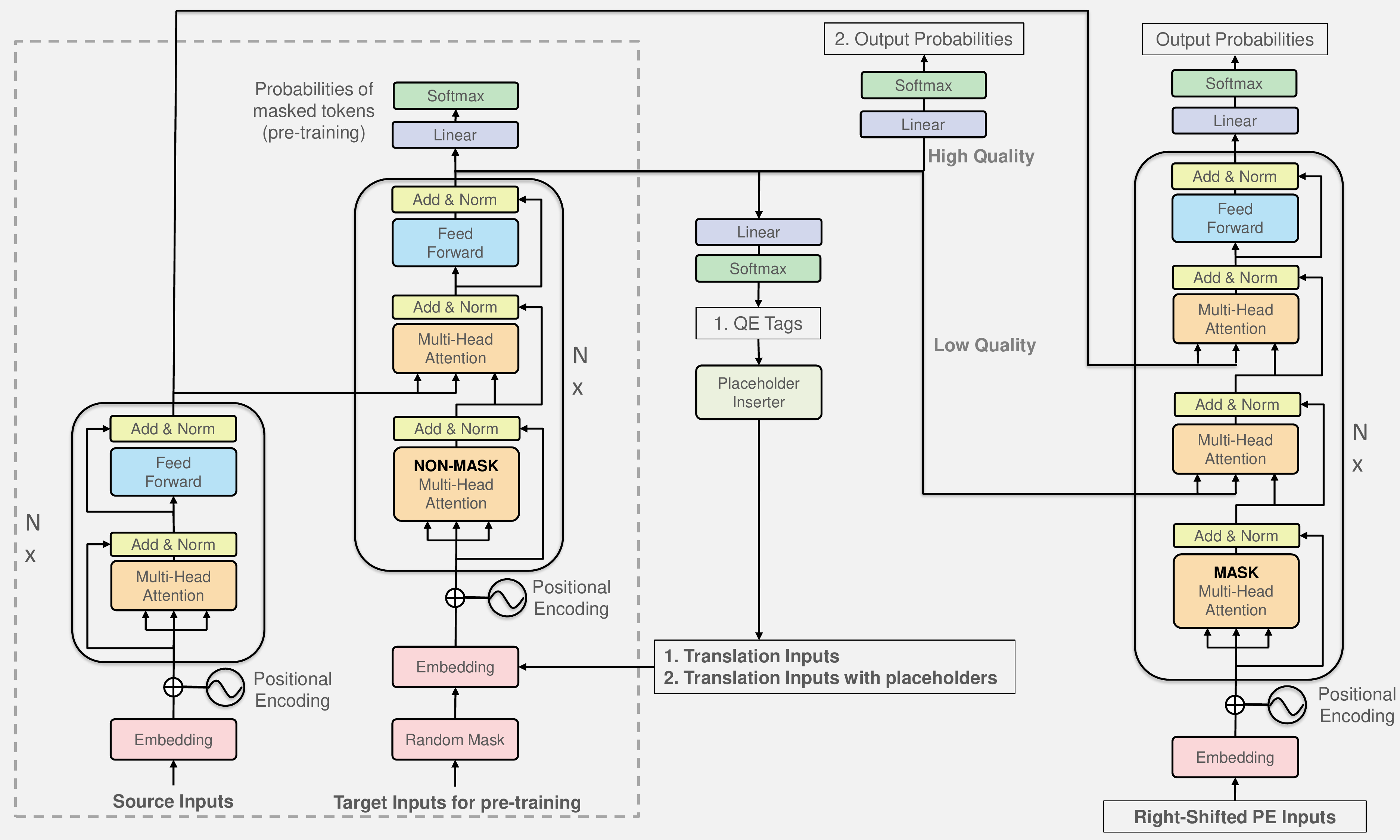}
  \caption{The detailed computational graph including detailed operations.}
  \label{fig:detail_model}
\end{figure*}
\subsection{Atomic Operation Automatic Post-Editing}
\label{sec:aoape}

\begin{figure}[t]
  \centering
  	\includegraphics[width=0.95\columnwidth]{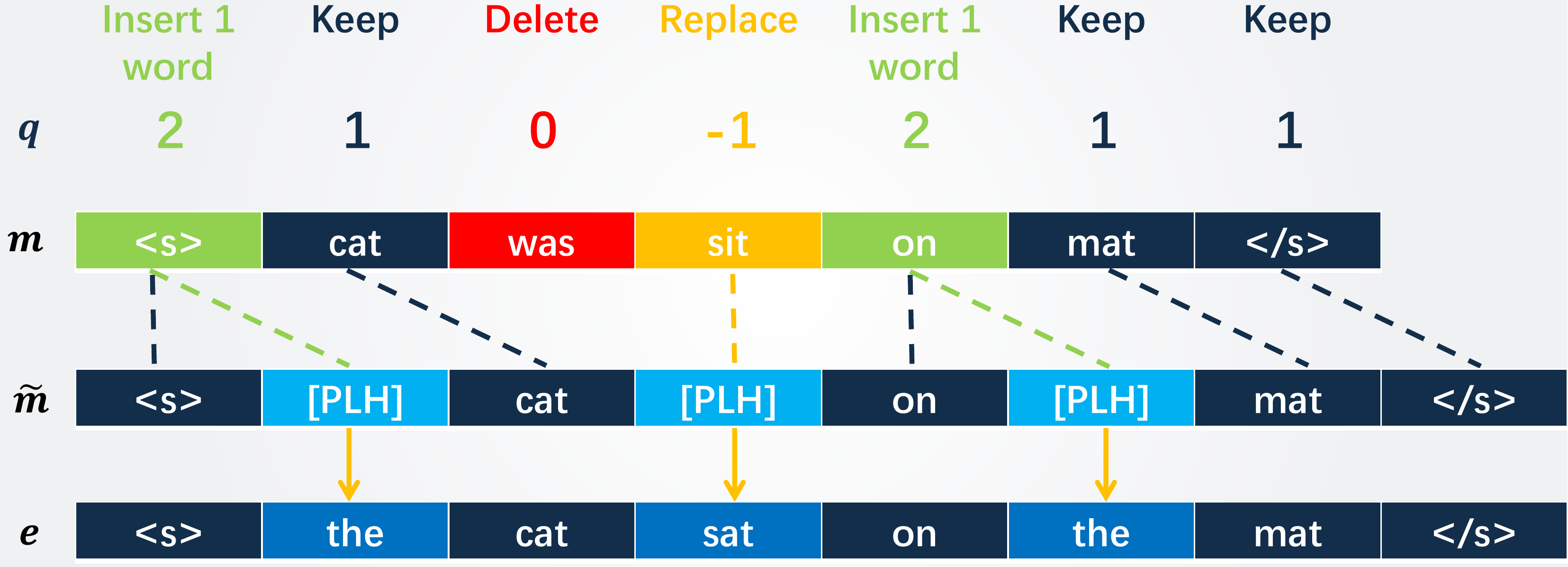}
    \caption{An example illustration of placeholder inserter and atomic operation APE.}
  \label{fig:plh_ins}
\end{figure}

The key idea of atomic operation APE is to reduce all predefined operations (insertion, deletion, substitution) into a special substitution operation by introducing an artificial token placeholder [PLH]. 

First, we align the machine translation $\mathbf{m}$ and the post-edits $\mathbf{e}$ by inserting [PLH]s, resulting in a new $\tilde{\mathbf{m}}$ of the same length as $\mathbf{e}$.  
Technically, we insert $q_i-1$ [PLH]s after $m_i$ if $q_i > 1$; we delete the current token $m_i$ if $q_i=0$; we replace $m_i$ with [PLH] if $q_i=-1$. 
For convenience, this process is denoted as $\tilde{\mathbf{m}}=\text{PLH\_INS}(\mathbf{m}, \mathbf{q})$. 

Second, the original APE task is transformed into another sequence tagging problem, since $|\tilde{\mathbf{m}}| = |\mathbf{e}|$.
\begin{equation}\label{eq:aoape}
\begin{split}
&P_{\text{PE}}^A(\mathbf{e}|\mathbf{s},\mathbf{m}) = P_{\text{PE}}^A(\mathbf{e}|\mathbf{s},\tilde{\mathbf{m}}) \\
=& \text{Softmax}_{\text{PE}}(\text{Enc}^M(\tilde{\mathbf{m}}, \text{Enc}(\mathbf{s})))
\end{split}
\end{equation}
Notice that \textbf{i)} the encoder and memory encoder share the parameters with the QE in Equation~(\ref{eq:qe_tag2}); \textbf{ii)} the softmax layer is different, because the number of outputs in APE has a different size equal to the vocabulary size. 
An intuitive visualization can see the Figure~\ref{fig:plh_ins} and the holistic pipeline sees the Figure~\ref{fig:model}. 

\subsection{Generative Automatic Post-Editing}
\label{sec:gam}

The larger HTER $h$ is, the lower quality of $\mathbf{m}$ is, and the more atomic operations are required. 
In this case, the previous APE model may be not powerful enough to learn complicated editing behaviors. 
We propose a backup APE model via auto-regressive approach for the deteriorated translations. 
Concretely, we write the dual-source language model into its probabilistic formulation.
\begin{align}\label{eq:gape}
&P_{\text{PE}}^G(\mathbf{e}|\mathbf{s},\mathbf{m}) = \prod_i P_{\text{PE}}^G(e_i|\mathbf{e}_{<i}, \mathbf{s},\mathbf{m}) \nonumber \\
=& \prod_i \text{Dec}(\mathbf{e}_{<i}; \text{Enc}^M(\mathbf{m}, \text{Enc}(\mathbf{s})); \text{Enc}(\mathbf{s}))
\end{align}
Notice that \textbf{i)} the encoder and memory encoder are still reused here, \textbf{ii)} the $\text{Dec}(\cdot;\cdot;\cdot)$ is a transformer decoder with hierarchical attention, since two memory blocks $\text{Enc}^M(\mathbf{m}, \text{Enc}(\mathbf{s}))$ and $\text{Enc}(\mathbf{s})$ are both conditional variables for the auto-regressive language model; \textbf{iii)} unlike sequence tagging, the inference of the generative APE is intrinsically non-parallelizable. 

\begin{algorithm}[t]
\caption{Imitation Learning Algorithm}
\begin{algorithmic}[1]
\small
\REQUIRE $\mathbf{s}$, $\mathbf{m}=\{m_i\}_{i=1}^M$, $\mathbf{e}=\{e_i\}_{i=1}^N$, hyperparameter $\beta\in(0,1)$.
\STATE Draw a random number $r$ from uniform distribution $[0,1]$.
\IF{$r > \beta$} \STATE $\tilde{\mathbf{m}} = \text{PLH\_INS}(\mathbf{m}, \mathbf{q})$.
\ELSE \STATE Randomly replace 20\% of $e_i$ as [PLH] to obtain $\tilde{\mathbf{m}}$.
\ENDIF
\STATE \textbf{Pseudo data for insertion} Remove all [PLH] in $\tilde{\mathbf{m}}$ to obtain $\mathbf{m}^i$.
\STATE \textbf{Pseudo data for substitution} Run APE inference model to obtain the prediction $\hat{\mathbf{e}}^s \leftarrow P_{\text{PE}}^A(\cdot|\mathbf{s}, \tilde{\mathbf{m}})$.
\STATE \textbf{Pseudo data for deletion} Randomly insert one or two [PLH]s to each gap in $\mathbf{e}$ with probability $0.15$ or $0.025$ to obtain the updated $\tilde{\mathbf{m}}$.
\STATE Run APE inference model to obtain the prediction\\
$\hat{\mathbf{e}}^d \leftarrow P_{\text{PE}}^A(\cdot|\mathbf{s}, \tilde{\mathbf{m}})$.
\RETURN 3 fake data points, $\mathbf{m}^i, \mathbf{m}^s = \hat{\mathbf{e}}^s, \mathbf{m}^d = \hat{\mathbf{e}}^d$.
\end{algorithmic} 
\label{alg:da}
\end{algorithm}

\subsection{Pre-training and Imitation Learning}
\label{sec:tricks}

Because of the lack of human post-editing data, training from scratch is typically difficult. 
We thus employ two workaround methods to improve the model performance. 

\textbf{Pre-training} It is worth noting that the reduced atomic operation APE is actually equivalent to the mask language modeling problem, a.k.a. the famous BERT \cite{devlin2019bert}. 
Therefore, we pre-train the encoder-memory encoder model as a conditional BERT with the data pairs $(\mathbf{s}, \mathbf{t})$ and $(\mathbf{m}, \hat{\mathbf{e}})$, aiming at learning the syntactic and alignment information of the ground truth. 
To make the pre-training valid on downstream tasks, we consistently use [PLH] token to randomly mask the reference / post-editing sentence. 

\textbf{Imitation Learning} As mentioned in \ref{sec:qe}, during inference, the predicted QE tags will causally tie to the successive APE algorithm, because $\tilde{\mathbf{m}}$ is derived from $(\mathbf{m}, \hat{\mathbf{q}})$. 
Although we would want the model to learn to predict all three atomic operations together, the small size of real post-editing data severely limits the performance of joint QE tagging. 
Therefore, we propose a model specialization strategy where the model learns three separate tasks: deletion, insertion, and substitution. 
A reasonable amount of training data can be generated for each of the tasks and the model learns to specialize in each operation. 
The details are summarized in Algorithm~\ref{alg:da}.

\begin{algorithm}[t]
\caption{APE Training}
\begin{algorithmic}[1]
\small
\REQUIRE Pre-training data $\mathcal{P}$ in pair $(\mathbf{s}, \mathbf{t} \text{ or } \mathbf{e})$, QE Training data $\mathcal{Q}$ in triplet$(\mathbf{s}, \mathbf{m}, \mathbf{e})$.
\STATE Pre-train the encoder-memory encoder model with $\mathcal{P}$ as \ref{sec:tricks}.
\WHILE{not converge}
	\STATE Sample a tuple from $\mathcal{Q}$.
	\STATE Call Algorithm~\ref{alg:da} to enlarge the training sample four times.
	\FOR{each $(\mathbf{s}, \mathbf{m}, \mathbf{e})$ in the augmented data}
		\STATE Calculate true QE tags $\mathbf{q}=$Algorithm~\ref{alg:qe_tag}$(\mathbf{m}, \mathbf{e})$.
		\STATE Get machine translation with [PLH]\\ $\tilde{\mathbf{m}} = \text{PLH\_INS}(\mathbf{m}, \mathbf{q})$.
		\STATE Update model parameters of encoder-memory encoder by optimizing the loss\\ $\mathcal{L}_{\text{QE}}(\mathbf{q},\mathbf{s},\mathbf{m}) + \mathcal{L}_{\text{PE}}^A(\mathbf{e},\mathbf{s},\tilde{\mathbf{m}})$.
		\STATE Update All model parameters by optimizing loss $\mathcal{L}_{\text{PE}}^G(\mathbf{e},\mathbf{s},\mathbf{m})$.
	\ENDFOR
\ENDWHILE
\RETURN All model parameters.
\end{algorithmic} 
\label{alg:train}
\end{algorithm}

\begin{algorithm}[t]
\caption{APE inference}
\begin{algorithmic}[1]
\small
\REQUIRE $\mathbf{s}$, $\mathbf{m}$, HTER threshold $\tau$, iteration steps $S$.
\STATE $\mathbf{m}^{(0)} = \mathbf{m}$
\FOR{$i=1,...,S$} 
	\STATE Run QE inference $\hat{\mathbf{q}}\leftarrow P_{\text{QE}}(\cdot|\mathbf{s}, \mathbf{m}^{(i-1)})$.
	\STATE Run Equation~{\ref{eq:hter}} to obtain quality metric $h$.
	\IF{$i==1$ and $h > \tau$}
		\STATE Run generative APE inference $\hat{\mathbf{e}}\leftarrow P_{\text{PE}}^G(\cdot|\mathbf{s}, \mathbf{m})$.
		\RETURN APE $\hat{\mathbf{e}}$.
	\ENDIF
	\STATE $\tilde{\mathbf{m}} = \text{PLH\_INS}(\mathbf{m}^{(i-1)}, \mathbf{q})$
	\STATE Run atomic operation APE inference\\
	$\mathbf{m}^{(i)}\leftarrow P_{\text{PE}}^A(\cdot|\mathbf{s}, \tilde{\mathbf{m}})$.
\ENDFOR
\RETURN APE $\hat{\mathbf{e}}=\mathbf{m}^{(S)}$.
\end{algorithmic} 
\label{alg:infer}
\end{algorithm}

\subsection{Training and Inference Algorithms}
\label{sec:alg}

In this section, we assemble all modules together into the final system.  
Because our model involves a nontrivial pipeline, we describe the details of training and inference separately and summarize them in Algorithm~\ref{alg:train} and \ref{alg:infer}. 

\textbf{Training} usually requires to minimize the loss function (negative data log-likelihood of probabilistic models) by stochastic gradient descent (SGD) with respect to the trainable parameters. 
Our QE and atomic operation APE are both sequence tagging task, while the generative APE is a sequence generation task. 
The three loss functions are uniformly defined as sequential cross entropy between the predicted and the true sequence. 
Note that the QE and atomic operation APE share the encoder-memory encoder, so these two losses can be summed together for optimization. 
However, the generative APE model has an isolated hierarchical transformer decoder, so we need a second update by optimizing the corresponding loss alone. 

\textbf{Inference} of our APE system is not quite the same as the training. 
First, the overall inference is a continuously alternating procedure between QE and APE, where the predicted APE is assigned as a new machine translation for iterative updating. 
However, the inner loop in training algorithm regards to the augmented data points. 
Second, we introduce an early stop after the first QE tagging prediction. 
If the predicted quality is very low ({\itshape i.e.} the HTER is larger than a cross-validated threshold), the generative APE will be called and the inference will immediately exit without further iterations. 
Lastly, the APE results are 
utilized by professional translators for further editing. 
In the next section, we validate the gain of APE over machine translation with regards to the efficiency.

\section{Experiments on our Proposed Model}
\label{sec:exp}

We verify the validity and efficiency of the proposed APE model by conducting a series of APE experiments and human evaluation on WMT'17 APE dataset. 
For convenience, we denote the generative post-editing model as {\itshape GM}, the atomic operation post-editing model as {\itshape AOM}, and the final hierarchical model as {\itshape HM} in this section.

\subsection{Setup}

\label{sec:setup}
\textbf{Dataset.} The open public WMT17 Automatic Post-Editing Shared Task \cite{bojar-etal-2017-findings} data on English-German (En-De) is widely used for APE experiments. 
It consists of 23K real triples (source, machine translation \& post-editing) for training and another 2K triples for testing from the Internet Technology (IT) domain. 
Besides, the shared task also provides a large-scale artificial synthetic corpus containing around 500K high quality and 4 million low quality synthetic triples. 
We over sample the APE real data by 20 times and merge it with the synthetic data, results in roughly 5 million of triples for both pre-training and APE training. 
The details of the training set are shown in Appendix Table \ref{tab:dataset details}. 
We adopt test set of the same task in WMT16 as the development set. Furthermore, we apply truecaser \cite{Koehn2007MosesOS} to all files and encode every sentence into subword units \cite{Kudo2018SentencePiece} with a 32K shared vocabulary.

\begin{table*}[tbp]
  \centering
  \caption{Performance Comparison on WMT17 APE En-De Dataset}
  \label{tab:overall results}
  \begin{small}
  \begin{tabularx}{\textwidth}{cccX}
    \toprule
    Model & BLEU$\uparrow$ & TER$\downarrow$ & Note \\
    \midrule
    Official Baseline & 62.49 & 24.48 & Do nothing with the origin machine translation \\
    MS-UEdin & 69.72 & 19.49 & Single model \cite{MSUEdin18APE}, winner of WMT18 APE task \\
    Levenshtein Transformer & 70.1 & 19.2 & Single model \cite{gu2019levenshtein} \\
    Unbabel & 70.66 & 19.03 & Single model \cite{unbabelSMT}, winner of WMT19 APE task. \\
    FBK (Ensemble) & 70.07 & 19.60 & Ensemble model\cite{FBK17APE}, winner of WMT17 APE task \\
    MS-UEdin (Ensemble) & 70.46 & 19.03 & Ensemble model\cite{MSUEdin18APE} \\
    Unbabel (Ensemble) & 71.90 & 18.07 & Ensemble model\cite{unbabelSMT}\\
    \midrule
    Only GM & 71.52 & 18.44 & Single model, {\itshape i.e.} $\tau=0$ in Algorithm~\ref{alg:infer} \\
    Only AOM & 68.40 & 20.34 & Single model, {\itshape i.e.} $\tau=1$ in Algorithm~\ref{alg:infer} \\
    Our HM & \textbf{72.07} & \textbf{18.01} & Single model, {\itshape i.e.} $\tau=0.3$, determined on development dataset \\
    \bottomrule
  \end{tabularx}
  \end{small}
\end{table*}

\begin{figure}[tbp]
  \centering
  	\includegraphics[width=0.95\columnwidth]{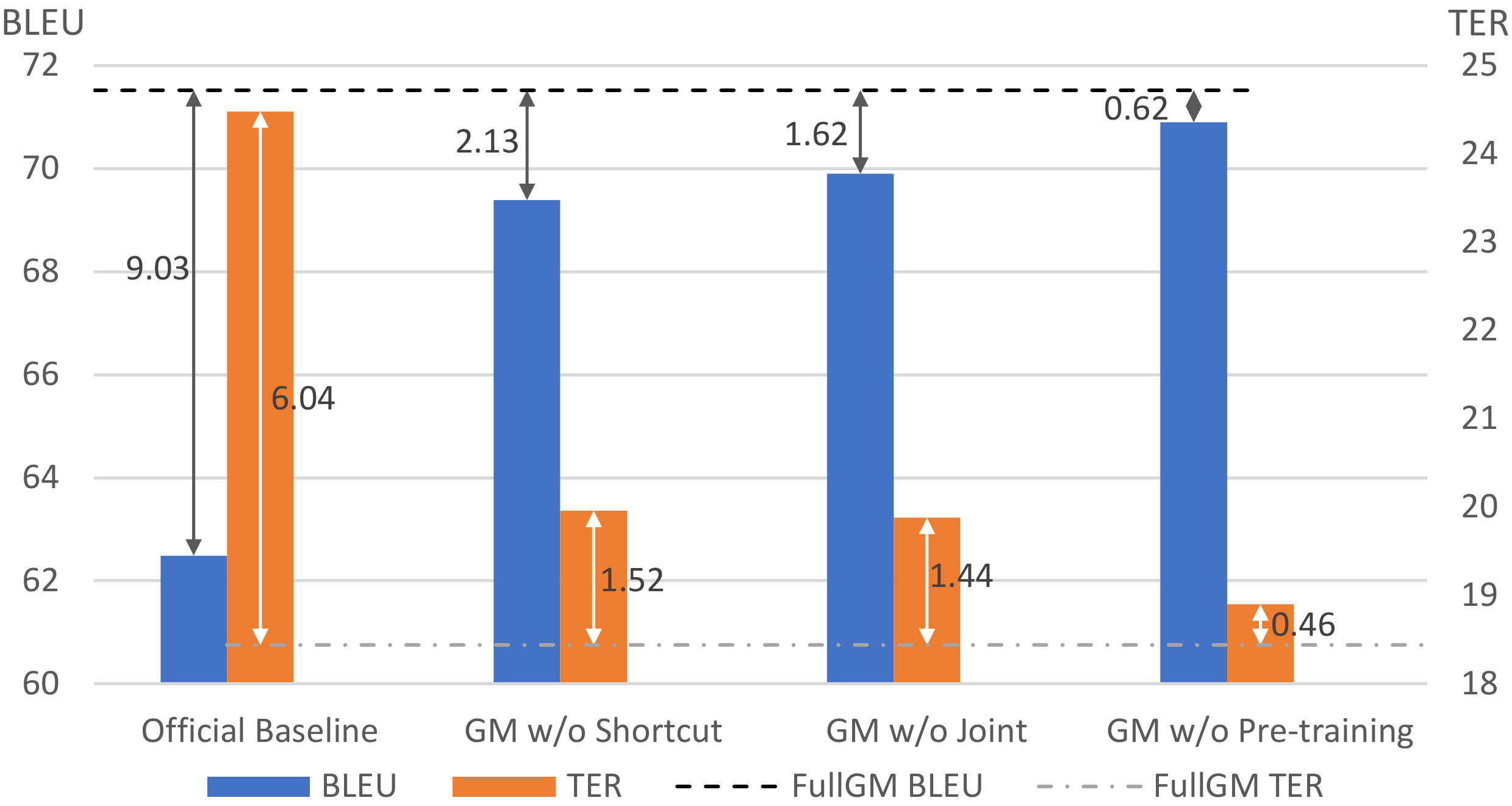}
    \caption{Results of Our Generative Model on Test Set}
  \label{fig:generative model results}
\end{figure}


\textbf{Evaluation Metrics.} We mainly evaluate our systems with metrics bilingual evaluation understudy (\textbf{BLEU}) \cite{Papineni2002BLEU} and translation edit rate (\textbf{TER}) \cite{Snover06TER}, since they are standard and widely employed in the APE shared task. 
The metric BLEU indicates how similar the candidate texts are to the reference texts, with values closer to 100 representing higher similarity.  
TER measures how many edits required from the predicted sentence to the ground truth sentence, and is calculated by Equation~(\ref{eq:hter}) as well and multiplied by 100.

\textbf{Training Details.} All experiments are trained on 8 NVIDIA P100 GPUs for maximum 100,000 steps for about two days until convergence, with a total batch-size of around 17,000 tokens per step and the Adam optimizer \cite{Kingma14Adam}. 
Only the source and post-edited sentence pairs are used for pre-training. 
During pre-training, 20\% tokens in post-editing sentence are masked as [PLH]. 
Parameters are being tuned with 12,000 steps of learning rates warm-up \cite{vaswani2017attention} for both of the GM and AOM model. 
However, 5 automatic post editing iterations ({\itshape i.e.} $S=5$ in Algorithm~{alg:infer}) are applied during the inference for the AOM model due to its characteristic of fine-grained editing behaviors. 
Except these modifications, we follow the default transformer-based configuration \cite{vaswani2017attention} for other hyper-parameters in our models. 

\subsection{APE Systems Comparison}
\label{sec:res}
The main results of automatic post-editing systems are presented in Table \ref{tab:overall results} and competitively compared with results of recent years' winners of WMT APE shared task and several other top results. 
It is observed that our hierarchical single model achieves the state-of-the-art performance on both BLEU and TER metrics, outperforming not only all other single models but also the ensemble models of top ranked systems in WMT APE tasks. 

Note that our hierarchical system is not a two-model ensemble.
The standard ensemble method requires inference and combination of results from more than one models.
In contrast, our hierarchical model contains multiple parameter-sharing modules to accomplish multi-tasks, and only need to infer once on the selected model. 

\subsubsection{Results of Generative APE Model}

As mentioned in section \ref{sec:gam}, the decoder of our generative model receives encoder-memory encoder outputs, refering to SRC memory and SRC-MT joint memory. 
A transformer attention layer encodes the SRC into the SRC memory, and the joint memory is produced by another one, which encodes the original MT conditionally on the SRC memory. 
These two encoders are pre-trained with sources and post-edits from the full training data.

We designed a set of systematic experiments to verify that our model benefits from such a design in Figure \ref{fig:generative model results}:
(1) To verify that the memory encoder has the ability to learn cross-lingual knowledge, we replace the memory encoder with an ordinary multi-head self-attention encoder, which does not accept the source memory as input, marked by {\itshape w/o Joint}. 
(2) To prove that the shortcut from the SRC memory to the decoder input is necessary, the shortcut is removed in the {\itshape w/o Shortcut} experiment. 
(3) To verify that our model can leverage representations from pre-training, we conduct an experiment without pre-training, denoted as {\itshape w/o Pre-training}.

%
%
The ablation results significantly demonstrate that our model does benefit from meory encoder, SRC memory shortcut and pre-training. 
Removing any of them will result in performance loss.

\subsubsection{Results of Atomic Operation APE Model}

\begin{figure}[!bh]

\centering

\includegraphics[width=0.9\columnwidth]{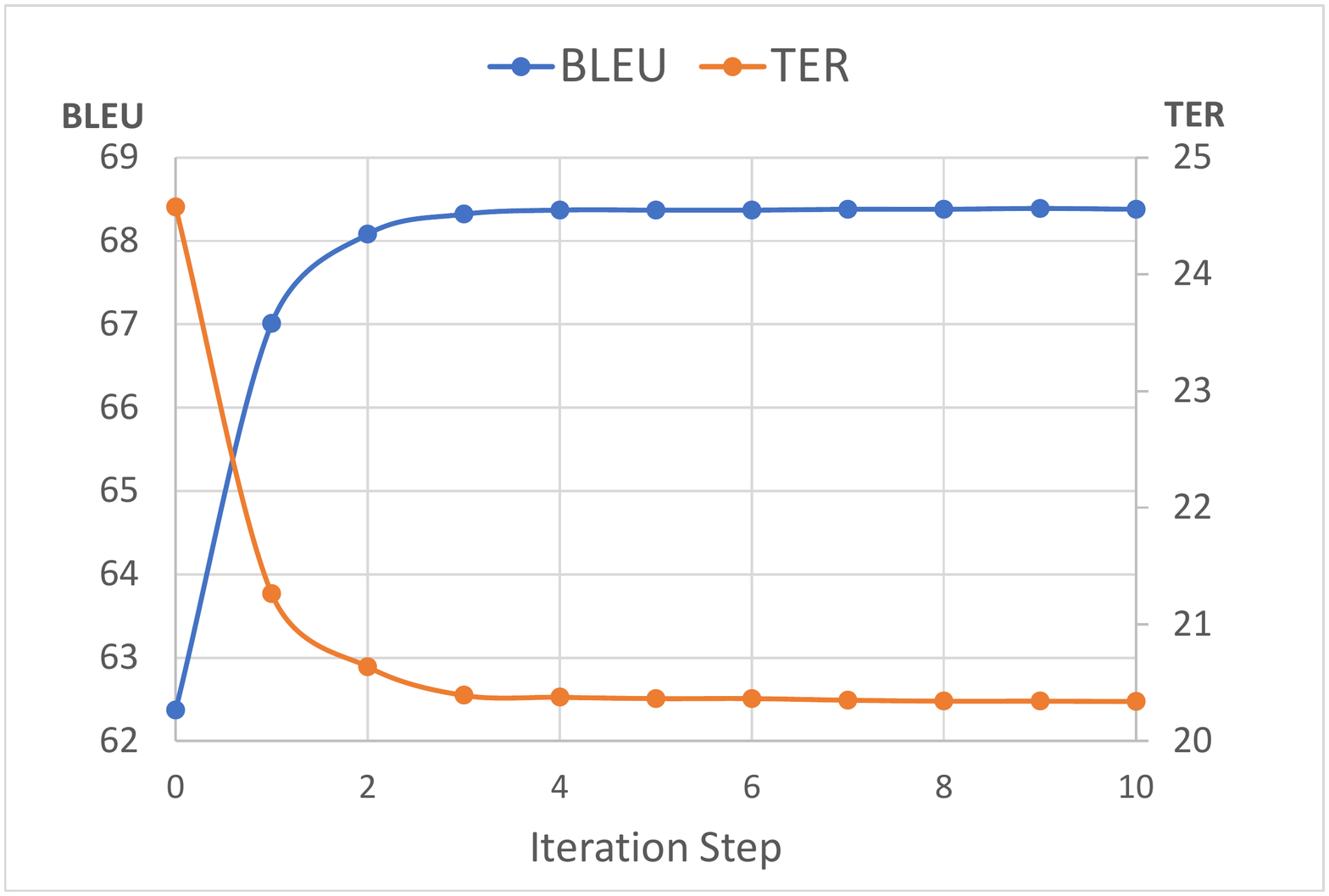}
\caption{The convergence curves of the AOM inference w.r.t. iteration. The iterative updating converges within only 3 to 5 steps, which is much smaller than the averaged number of decoding steps of the GM.}
\label{fig:IterResult}
\end{figure}

\begin{table}[tbp]
\setlength{\tabcolsep}{2pt}
  \centering
  \caption{Performance Gain from Pseudo Data}
  \label{tab:pseudo data result}
  \begin{small}
  \begin{tabularx}{\columnwidth}{Xcccc}
    \toprule
    Model & BLEU$\uparrow$ & TER$\downarrow$ & $\Delta$BLEU & $\Delta$TER\\
    \midrule
    AOM w/o pseudo data & 65.65 & 22.14 & - & - \\
    AOM with pseudo data & 68.40 & 20.34 & \textbf{$+$2.75} & \textbf{$-$1.80}\\
    \bottomrule
  \end{tabularx}
  \end{small}
\end{table}
In each iteration, based on the QE model's output, our AOM refines the MT in parallel regarding to all placeholders.
Unlike the GM, the time cost of the AOM only depends on the steps of iterations, regardless of the length of the sentence.
To evaluate the decoding efficiency, we collect the AOM's performances at different iteration steps, as shown in Figure \ref{fig:IterResult}. 

\textbf{The Role of Pseudo Data.} As noted in section \ref{sec:tricks}, model specialization algorithm is applied to train the model to learn different kinds of atomic operations. 
We compare our AOM on the test set with and without pseudo data in Table \ref{tab:pseudo data result}. 
The results demostrate that our model specialization algorithm plays a key role by providing a powerful guidance for training and making up for the deficiency from the lack of large amount of real APE data.

\subsubsection{Results of QE Model}

The QE model is the prerequisite of the final hierarchical model as well as the basis of our atomic operation model. 
Therefore, it is necessary to guarantee the performance of QE results as accurate as possible. 
Unlike the traditional OK/BAD word-level QE task in WMT \cite{bojar-etal-2017-findings}, our model pursues to predict fine-grained quality tags. So, we cannot make a completely fair comparison with previous works. 

The fine-grained quality tag of each word predicted by the model can be classified into one of the four labels: {\itshape K} for \textbf{Kept}, {\itshape E} for \textbf{Erroneous
}, {\itshape R} for \textbf{Redundant} and {\itshape M} for \textbf{Missing}. 
Furthermore, we convert the predicted fine-grained QE tags to OK/BAD tags directly by treating tag {\itshape K} and tag {\itshape M} as {\itshape OK}, and the other two tags as {\itshape BAD} according to the rules of tagging in WMT17 QE Shared Task. 

We provide our fine-grained QE results on the test dataset of WMT17 APE Task in Table \ref{tab:QE results}, where the ground-truth tags are produced by Algorithm~\ref{alg:qe_tag} in Appendix \ref{appendix:levt}. 
Note that the TER score can be easily computed from the predicted quality tags. 
The predicted TER score is regarded as an indicator of MT quality in our hierarchical model: MTs with quality higher than $\tau$ in Algorithm~\ref{alg:infer} are fed to the GM, otherwise they are sent to the AOM. 
The hyper-parameter $\tau=0.3$ is determined by cross validation on WMT16 development dataset. 
Afterwards, we apply it on the WMT17 test dataset to select a potentially preferable model from GM and AOM to generate the final APE result for each SRC and MT pair.

There are more than $75\%$ of tokens in the training set are tagged with {\itshape Keep}. 
In terms of the huge challenge posed by the unbalanced dataset, our fine-grained quality estimation is quite remarkable. 
The performance of our final hierarchical model in Table \ref{tab:overall results} proves the effectiveness of it.

\begin{table}[t]
\setlength{\tabcolsep}{3pt}
  \centering
  \caption{Results of Fine-Grained QE Model (Pearson = 0.664). Quality tag prediction is evaluated in terms of multi-classification accuracy via F1-scores. The overall MT quality estimation is measured by the Pearson correlation coefficient, indicating the correlation between the predicted and the real MT quality w.r.t. TER.}
  \label{tab:QE results}
  \begin{small}
  \begin{tabular}{ccccccc}
    \toprule
     & K & E & R & M & OK & BAD \\
    \midrule
    Precision$\uparrow$ & 0.877 & 0.710 & 0.563 & 0.622 & 0.898 & 0.783 \\
    Recall$\uparrow$ & 0.951 & 0.471 & 0.480 & 0.540 & 0.962 & 0.559 \\
    F1-score$\uparrow$ & 0.913 & 0.566 & 0.518 & 0.578 & 0.928 & 0.652 \\
    \bottomrule
  \end{tabular}
  \end{small}
\end{table}

\subsection{Results of Human Evaluation}
\label{exp:human_eval}
\begin{figure}[!bht]
  \centering
  	\includegraphics[width=0.95\columnwidth]{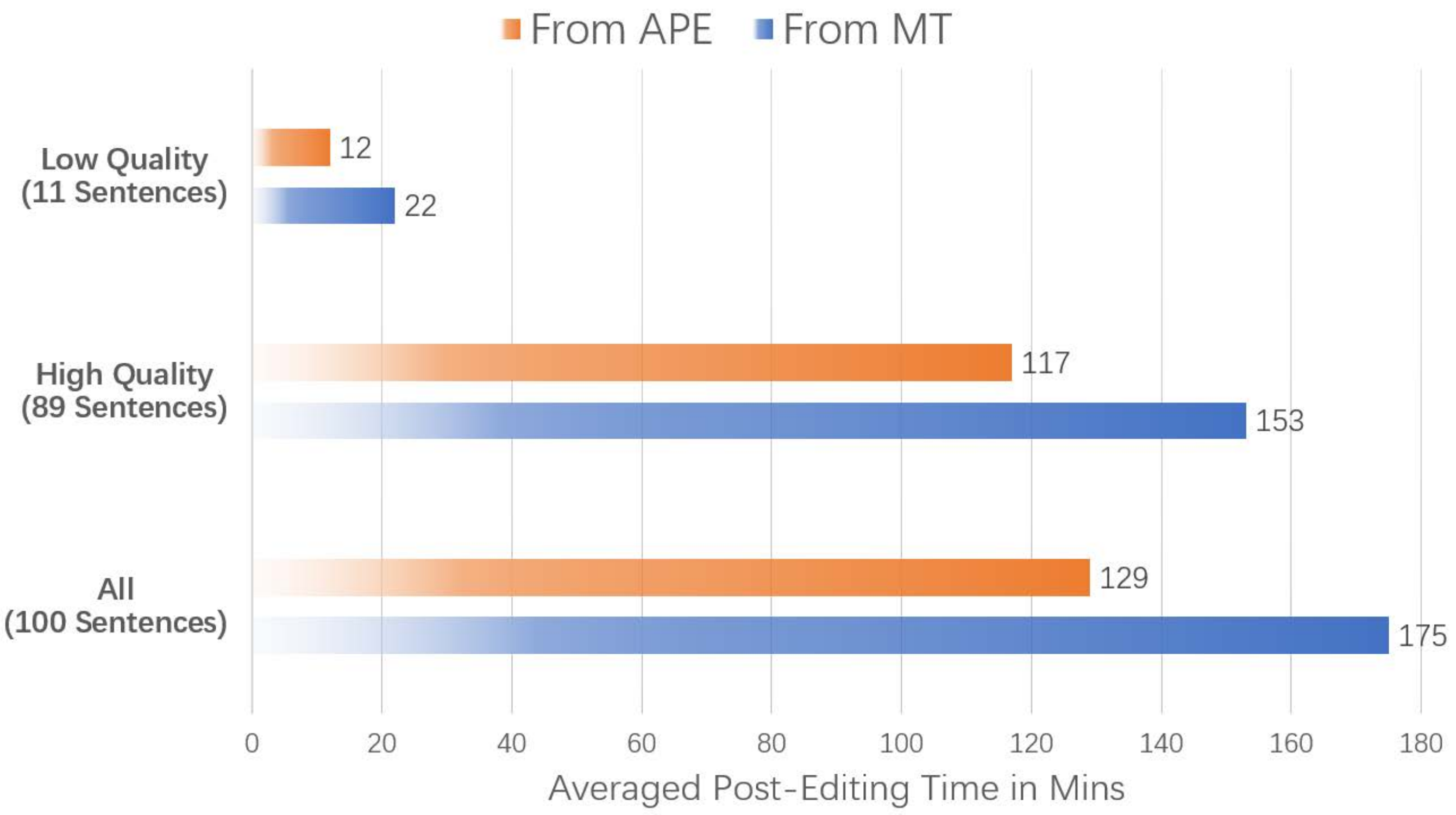}
    \caption{Time Spent in Post-Editing by Translators. 
    The averaged total time spent by translators to post-edit the APE becomes significantly decreased by 26.3\%
    }
  \label{fig:case_study_2}
\end{figure}

We conduct real post-editing experiments with professional translators involved. 
There are 6 independent participating translators, randomly divided into 2 groups. They are all native speakers of German and have 10+ years of experience in translation of En-De in IT related domains. 
We follow two different flows in our experiments. 
For fair comparison, both of the two groups see the same 100 source sentences picked from the WMT17 test dataset. 
The MTs are provided for the first group for post-editing, while our model generated APEs for the second group. 
However, the information on the category of the translation is not revealed to translators. 
The translators are asked to record the elapsed time of their labor in total.

The statistics of averaged post-editing time for different translators are summarized in Figure~\ref{fig:case_study_2}. 
Besides the total time, we also analyze the duration for low and high quality translations separately (determined by QE model). 
In either case, post-editing from the APE costs less time. We also did case study about high-quality vs low-quality APE in Appendix \ref{appendix:human_eval}. 
From different perspectives of experimental validation, we can conclude that the APE generated by our model can ease the burden of translators and substantially improve the post-editing efficiency.

\section{Conclusion}
In this paper, we proposed a hierarchical model that utilizes the fine-grained word-level QE prediction to select one of the two APE models we proposed to generate better translations automatically, which shows a state-of-the-art performance.  
In particular, we designed a dynamic deep learning model using imitation learning, which intuitively mimics the editing behaviors of human translators. 
Our hierarchical model is not a standard ensemble model in the conventional sense. We merely shared the parameters of different modules to accomplish different objectives, including QE, AOM and GM. 
Our experimental findings show that if the characteristics of errors in the machine translation can be accurately simulated, it is highly likely that MT output can be automatically refined by the APE model. 
Towards this end, we conduct a rigorous comparison of the machine translation and automatic post-editing based manual post-editing tasks, and it is observed that the latter can significantly increase the efficiency of post-editing. 

\section*{Acknowledgments}

This work is partly supported by National Key R\&D Program of China (2018YFB1403202).


\bibliographystyle{acl_natbib}
\bibliography{anthology,emnlp2020}

\cleardoublepage
\appendix

\section{Appendix}

\subsection{Pseudo code of QE tag computation}
\label{appendix:levt}
The computation of QE tags is quite similar to the famous Minimum Edit Distance problem and can be solved with dynamic programming in algorithm \ref{alg:qe_tag}.
\begin{algorithm}[!hbt]
\caption{QE tag computation}
\begin{algorithmic}[1]
\small
\REQUIRE machine translation $\mathbf{m}=\{m_i\}_{i=1}^M$, post-editing $\mathbf{e}=\{e_i\}_{i=1}^N$.
\STATE Initialize the edit distance matrix $d_{i,0}=i, d_{0,j}=j$ and QE tag $q_i=1$.
\FOR{$i=1,...,M$}
	\FOR{$j=1,...,N$}
		\STATE $d_{i,j} = \min\{d_{i-1,j-1} + \mathbb{I}_{m_i\neq e_j}, d_{i,j-1} + 1, d_{i-1,j} + 1\}$
	\ENDFOR
\ENDFOR
\WHILE{$i>0$ or $j>0$}
	\IF{$i>0$ and $j>0$ and $d_{i-1,j-1} + 1 = d_{i,j}$}	\STATE $q_i=-1$, $i--$, $j--$
	\ELSIF{$j>0$ and $d_{i,j-1} + 1=d_{i,j}$}	\STATE $q_i++$, $j--$
	\ELSIF{$i>0$ and $d_{i-1,j} + 1=d_{i,j}$}	\STATE $q_i=0$, $i--$
	\ELSE \STATE $i--$, $j--$
	\ENDIF
\ENDWHILE
\RETURN $\mathbf{q}=\{q_i\}_{i=1}^M$
\end{algorithmic} 
\label{alg:qe_tag}
\end{algorithm}


\subsection{Details of the Traning Corpus}
WMT APE shared-task provided both real APE triplets and a large a large-scale artificial synthetic corpus containing around 500K high quality and 4 million low quality synthetic triples. Table \ref{tab:dataset details} shows the difference between them.

\begin{table}[!h]
\setlength{\tabcolsep}{2pt}
  \centering
  \caption{Details of the WMT 2017 APE Shared-Task Dataset. The BLEU and TER metrics are directly evaluated on machine translation and post-editings as references.}
  \label{tab:dataset details}
  \begin{small}
  \begin{tabular}{ccccc}
    \toprule
    Source & \# Sentence & Avg. Length & BLEU & TER\\
    \midrule
    Real Triples & 23,000 & 17.88 & 61.87 & 25.35 \\
    Artificial 500K & 526,368 & 20.90 & 60.01 & 25.55 \\
    Artificial 4M & 4,391,180 & 16.68 & 46.59 & 35.37 \\
    \midrule
    500K$+$20$*$Real & 986,368 & 19.49 & 60.80 & 25.46 \\
    4M$+$500K$+$20$*$Real & \multirow{2}{*}{5,377,548} & \multirow{2}{*}{17.20} & \multirow{2}{*}{49.65} & \multirow{2}{*}{33.31} \\
    (Full Training data)  &  &  &  &  \\
    \bottomrule
  \end{tabular}
  \end{small}
\end{table}

\subsection{Case Study and Runtime Efficiency}
\label{appendix:human_eval}
As mentioned in the paper, the AOM is more suitable for translations that only require a few edit operations while GM is more preferable for low quality translations. 
To demonstrate this conclusion and prove the effectiveness of our QE-based automatic selector, some cases of translations with different qualities are shown in Table \ref{tab:case_study_1}. 

In case 1 and case 2, the translation is quite close to {\itshape pe}. Therefore, the AOM only need to predict tokens for a small number of [PLH]s. 
When there are relatively complete contexts provided, the AOM can achieve a higher performance than the GM. 
Moreover, after reading the source and the final output, the human translators did not even take any additional action to improve the translation quality. 

In the opposite way, as shown in case 3 and case 4, there is a huge gap between {\itshape mt} and {\itshape pe}, and the input for AOM contains a considerable number of placeholders, which lacks enough contextual information. 
In these cases, our GM can auto-regressively regenerate the translation based on the given {\itshape mt} to guarantee the higher quality of the final output. 
Based on the QE selector, the translators only need to make very few efforts to correct the errors in the final generated APE of our model. 

\begin{table*}[htbp]
\scriptsize
  \centering
  \caption{Examples of Crowdsourcing after APE. Tokens in ``$\langle \rangle$" indicates GM's over corrections or AOM's inaccurate translations due to too many consecutive [PLH] predictions, which leads inadequate contextual information. Tokens in ``\{\}" highlights correct automatic editings. }
  \label{tab:case_study_1}
      \begin{tabularx}{\textwidth}{clX}
  \toprule
  \multicolumn{3}{c}{\textbf{High Quality s Translation Case}}\\
    \midrule
\multirow{11}*{Case1} & SRC & In List view , click any column header to sort by that criteria . \\ 
 & MT & Klicken Sie in der Listenansicht auf eine beliebige Spaltenüberschrift , um nach dieser Kriterien sortieren . \\ 
 & PE & Klicken Sie in der Listenansicht auf eine beliebige Spaltenüberschrift , um nach diesen Kriterien zu sortieren . \\  
 & MT (sub-word) & \_\_klicken \_\_Sie \_\_in \_\_der \_\_Listenansicht \_\_auf \_\_eine \_\_beliebige \_\_Spalten überschrift \_\_, \_\_um \_\_nach \_\_dieser \_\_Kriterien \_\_sortieren \_\_. \\ 
 & Predicted QE Tag & 1 1 1 1 1 1 1 1 1 1 1 1 1 -1 2 1 1 \\ 
 & TER vs Predicted TER & 11.76 vs 11.11 \\ 
 & AOM Input & \_\_klicken \_\_Sie \_\_in \_\_der \_\_Listenansicht \_\_auf \_\_eine \_\_beliebige \_\_Spalten überschrift \_\_, \_\_um \_\_nach [PLH] \_\_Kriterien [PLH] \_\_sortieren \_\_. \\ 
 & AOM Output & \_\_klicken \_\_Sie \_\_in \_\_der \_\_Listenansicht \_\_auf \_\_eine \_\_beliebige \_\_Spalten überschrift \_\_, \_\_um \_\_nach \textbf{\{\_\_diesen\}} \_\_Kriterien \textbf{\{\_\_zu\}} \_\_sortieren \_\_. \\ 
 & GM Output & \_\_klicken \_\_Sie \_\_in \_\_der \_\_Listenansicht \_\_auf \_\_eine \_\_beliebige \_\_Spalten überschrift \_\_, \_\_um \_\_nach \_\_dieser \_\_Kriterien \textbf{\{\_\_zu\}} \_\_sortieren \_\_. \\ 
 & Final Output & Klicken Sie in der Listenansicht auf eine beliebige Spaltenüberschrift , um nach \textbf{diesen} Kriterien \textbf{zu} sortieren . \\ 
 & Translator Edit & no action \\
 
 \midrule
\multirow{11}*{Case2} & SRC & You can justify all text in a paragraph either including or excluding the last line . \\ 
 & MT & Sie können den gesamten Text eines Absatzes mit oder ohne die letzte Zeile . \\ 
 & PE & Sie können den gesamten Text eines Absatzes mit oder ohne die letzte Zeile ausrichten . \\ 
 & MT (sub-word) & \_\_Sie \_\_können \_\_den \_\_gesamten \_\_Text \_\_eines \_\_Absatzes \_\_mit \_\_oder \_\_ohne \_\_die \_\_letzte \_\_Zeile \_\_. \\ 
 & Predicted QE Tag & 1 1 1 1 1 1 1 1 1 1 1 1 2 1 \\ 
 & TER vs Predicted TER & 6.67 vs 6.67 \\ 
 & AOM Input & \_\_Sie \_\_können \_\_den \_\_gesamten \_\_Text \_\_eines \_\_Absatzes \_\_mit \_\_oder \_\_ohne \_\_die \_\_letzte \_\_Zeile [PLH] \_\_. \\ 
 & AOM Output & \_\_Sie \_\_können \_\_den \_\_gesamten \_\_Text \_\_eines \_\_Absatzes \_\_mit \_\_oder \_\_ohne \_\_die \_\_letzte \_\_Zeile \textbf{\{\_\_ausrichten\}} \_\_. \\ 
 & GM Output & \_\_Sie \_\_können \_\_den \_\_gesamten \_\_Text \_\_eines \_\_Absatzes \textbf{$\langle$\_\_entweder \_\_einschließlich$\rangle$} \_\_oder \_\_ohne \_\_die \_\_letzte \_\_Zeile \_\_löschen \_\_. \\ 
 & Final Output & Sie können den gesamten Text eines Absatzes mit oder ohne die letzte Zeile \textbf{ausrichten} . \\
 & Translator Edit & no action \\
\cmidrule(lr){1-3}\morecmidrules\cmidrule(lr){1-3}
  
  \multicolumn{3}{c}{\textbf{Low Quality Translation Case}}\\
  \midrule
\multirow{11}*{Case3} & SRC & In Start Number , enter the number to assign to the first PDF on the list . \\ 
 & MT & Wählen Sie unter `` Number , " geben Sie die Nummer für die erste PDF-Datei in der Liste aus . \\ 
 & PE & Geben Sie unter `` Startnummer " die Nummer für die erste PDF-Datei in der Liste ein . \\  
 & MT (sub-word) & \_\_wählen \_\_Sie \_\_unter \_\_`` \_\_Number \_\_, \_\_" \_\_geben \_\_Sie \_\_die \_\_Nummer \_\_für \_\_die \_\_erste \_\_PDF - Datei \_\_in \_\_der \_\_Liste \_\_aus \_\_. \\ 
 & Predicted QE Tag & -1 1 1 2 -1 -1 -1 -1 -1 1 1 0 -1 -1 1 1 1 -1 1 1 -1 1 \\ 
 & TER vs Predicted TER & 35.29 vs 54.55 \\  
 & AOM Input & [PLH] \_\_Sie \_\_unter \_\_`` [PLH] [PLH] [PLH] [PLH] [PLH] [PLH] \_\_die \_\_Nummer [PLH] [PLH] \_\_PDF - Datei [PLH] \_\_der \_\_Liste [PLH] \_\_. \\ 
 & AOM Output & \textbf{\{\_\_geben\}} \_\_Sie \_\_unter \_\_`` \_\_Start \textbf{$\langle$\_\_geben \_\_Sie \_\_zum \_\_Zuweisen \_\_$\rangle$}" \_\_die \_\_Nummer \_\_der \_\_ersten \_\_PDF - Datei \_\_über \_\_der \_\_Liste \textbf{\{\_\_ein\}} \_\_. \\ 
 & GM Output & \textbf{\{\_\_geben\}} \_\_Sie \_\_unter \_\_`` \textbf{\{\_\_Start nummer\}} \_\_" \_\_die \_\_Nummer \_\_für \_\_die \_\_erste \_\_PDF - Datei \_\_in \_\_der \_\_Liste \_\_an \_\_. \\ 
 & Final Output & \textbf{Geben} Sie unter `` \textbf{Startnummer} " die Nummer für die erste PDF-Datei in der Liste an . \\ 
 & Translator Edit & an$\rightarrow$ein  \\
 
   \midrule
\multirow{11}*{Case4} & SRC & The Illustrator text is converted to HTML text with basic formatting attributes in the resulting web page . \\ 
 & MT & Die Illustrator Text HTML-Text mit grundlegenden Formatierungsattribute in der erstellten Webseite konvertiert wird . \\ 
 & PE & Die Illustrator-Text wird in HTML-Text mit grundlegenden Formatierungsattributen in der erstellten Webseite konvertiert . \\ 
 & MT (sub-word) & \_\_die \_\_Illustrator \_\_Text \_\_HTML - Text \_\_mit \_\_grundlegenden \_\_Formatierung s attribute \_\_in \_\_der \_\_erstellten \_\_Webseite \_\_konvertiert \_\_wird \_\_. \\
 & Predicted QE Tag & -1 3 3 1 1 1 1 1 1 1 -1 1 1 1 1 1 0 1 \\ 
 & TER vs Predicted TER & 35.29 vs 33.33 \\  
 & AOM Input & [PLH] \_\_Illustrator [PLH] [PLH] \_\_Text [PLH] [PLH] \_\_HTML - Text \_\_mit \_\_grundlegenden \_\_Formatierung s [PLH] \_\_in \_\_der \_\_erstellten \_\_Webseite \_\_konvertiert \_\_. \\ 
 & AOM Output & \_\_in \_\_Illustrator - Der \_\_Text \_\_in \_\_in \_\_HTML - Text \_\_mit \_\_grundlegenden \_\_Formatierung s \textbf{\{attributen\}} \_\_in \_\_der \_\_erstellten \_\_Webseite \_\_konvertiert \_\_. \\ 
 & GM Output & \_\_der \_\_Illustrator \textbf{\{- Text \_\_wird \_\_in\}} \_\_HTML - Text \_\_mit \_\_grundlegenden \_\_Formatierung s \textbf{\{attributen\}} \_\_in \_\_der \_\_erstellten \_\_Webseite \_\_konvertiert \_\_. \\ 
 & Final Output & Der Illustrator-\textbf{Text} \textbf{wird} \textbf{in} HTML-Text mit grundlegenden \textbf{Formatierungsattributen} in der erstellten Webseite konvertiert . \\ 
 & Translator Edit & Der$\rightarrow$Die \\

    \bottomrule
  \end{tabularx}
\end{table*}
A practical point of the computer assisted translation via APE is its expense and computational cost. 
Compared with the traditional computer assisted translation crowdsourcing, machine translation + human post-editing, our additional automatic post-editing does increase the computational cost, which is roughly equivalent to another machine translation model. 
In general, the crowdsourcing is charged by hours. 
The numbers in our findings suggest a promising budget cut associated with CAT crowdsourcing. 
However, this extra APE module may lead to a latency increase by \~400ms, which is still far below the average time cost by human post-editing. 
Even for an online crowdsourcing system, a well-designed concurrent mechanism should make the translators not feel any delay. 
From the perspective of architecture scale, the APE model can be deployed in the identical processing unit for the machine translation model and be called successively in a pipeline. 
The only concern is that the memory storage capacity should be large enough to store more parameters.

\end{document}